\documentclass[journal]{IEEEtran}
\usepackage{times}
\usepackage{epsfig}
\usepackage{graphicx}
\usepackage{amsmath}
\usepackage{amssymb}
\usepackage{algorithm}
\usepackage{algpseudocode}
\algrenewcommand\algorithmicindent{.9em}%
\usepackage{enumerate}
\usepackage{multirow}
\usepackage{enumitem}
\usepackage{makecell}
\usepackage{tabulary}
\usepackage{pifont}
\usepackage{boldline}
\usepackage[colorlinks,urlcolor=black,linkcolor=black,anchorcolor=black,citecolor=black]{hyperref}

\ifCLASSINFOpdf
\else
\fi

\begin{document}
%
\title{Single Shot Video Object Detector}
\author{Jiajun~Deng,~Yingwei~Pan,
        ~Ting~Yao,~\IEEEmembership{Member,~IEEE},
        ~Wengang~Zhou,~\IEEEmembership{Member,~IEEE},\\
        ~Houqiang~Li,~\IEEEmembership{Senior Member,~IEEE},
        ~and~Tao~Mei,~\IEEEmembership{Fellow,~IEEE}
\thanks{T. Yao and W. Zhou are the corresponding authors.}
\thanks{This work was supported in part to Dr. Houqiang Li by NSFC under contract No. 61836011, and in part to Dr. Wengang Zhou by NSFC under contract No. 61822208 \& 61632019 and Youth Innovation Promotion Association CAS (No. 2018497).}
\thanks{J. Deng, W. Zhou and H. Li are with University of Science and Technology of China, Hefei, China (e-mail: dengjj@mail.ustc.edu.cn; zhwg@ustc.edu.cn; lihq@ustc.edu.cn).}
\thanks{Y. Pan, T. Yao and T. Mei are with JD AI Research, Beijing, China (e-mail: panyw.ustc@gmail.com; tingyao.ustc@gmail.com; tmei@jd.com).}}
%

%



\maketitle

\begin{abstract}
Single shot detectors that are potentially faster and simpler than two-stage detectors tend to be more applicable to object detection in videos. Nevertheless, the extension of such object detectors from image to video is not trivial especially when appearance deterioration exists in videos, \emph{e.g.}, motion blur or occlusion. A valid question is how to explore temporal coherence across frames for boosting detection. In this paper, we propose to address the problem by enhancing per-frame features through aggregation of neighboring frames. Specifically, we present Single Shot Video Object Detector (SSVD) --- a new architecture that novelly integrates feature aggregation into a one-stage detector for object detection in videos. Technically, SSVD takes Feature Pyramid Network (FPN) as backbone network to produce multi-scale features. Unlike the existing feature aggregation methods, SSVD, on one hand, estimates the motion and aggregates the nearby features along the motion path, and on the other, hallucinates features by directly sampling features from the adjacent frames in a two-stream structure. Extensive experiments are conducted on ImageNet VID dataset, and competitive results are reported when comparing to state-of-the-art approaches. More remarkably, for $448 \times 448$ input, SSVD achieves 79.2\% mAP on ImageNet VID, by processing one frame in 85 ms on an Nvidia Titan X Pascal~GPU. The code is available at \url{https://github.com/ddjiajun/SSVD}.
\end{abstract}


%
\IEEEpeerreviewmaketitle

\section{Introduction}
\IEEEPARstart{T}he development of deep learning technologies has led to the significant surge of research activities in computer vision area. In between, object detection is one of the most fundamental tasks and the recent advances in deep convolutional neural networks \cite{he2016resnet,huang2017densely,krizhevsky2012imagenet,Simonyan:ICLR15,Szegedy:CVPR15} have successfully achieved remarkable improvements on object detection in images \cite{Cai_2018_CVPR,cai2020weighting,dai2016r,he2017mask,hu2018relation,lin2017focal,liu2016ssd,ren2015faster,Zhang_2018_CVPR}. Nevertheless, directly applying these object detectors for still images to object detection in videos is very challenging due to the fact that video is an information-intensive media with large variations and complexities, not to mention that some frames in videos may be deteriorated by motion blur or occlusion. Such facts motivate and highlight the explorations of object detectors in videos to improve detection accuracy. Furthermore, the natural existence of spatio-temporal coherence in videos also offers a fertile ground for designing video-level object detectors.

\begin{figure}[!tb]
\centering {\includegraphics[width=0.5\textwidth]{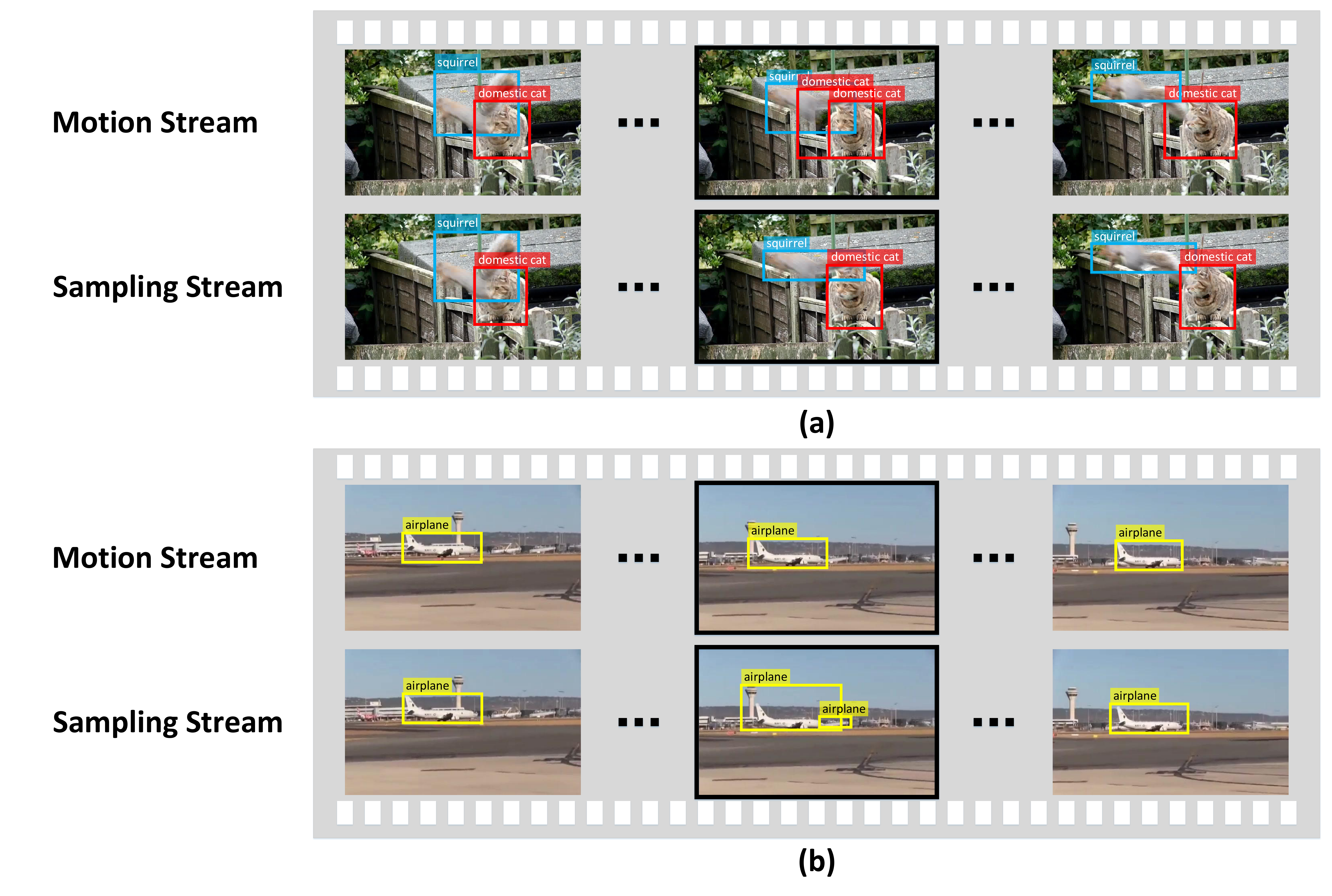}}
\vspace{-0.3in}
\caption{\small Video object detection results with motion stream and sampling stream. In the upper example (a), motion stream fails to detect domestic cat accurately due to the motion blur, while sampling stream that performs feature aggregation without motion estimation can ameliorate this case. In the bottom example (b) when the airplane moves fast, an unwanted bounding box of airplane is generated for sampling stream. In contrast, motion stream can solve this problem by capturing long-range motion with explicit motion estimation. Object detection in videos therefore should take both motion and sampling streams into consideration.
}
\label{fig.fig1}
\vspace{-0.2in}
\end{figure}

There are two general directions along the exploitation of spatio-temporal coherence for object detection in videos. One common solution for video object detection is box-level tracking \cite{feichtenhofer2017detect,han2016seq,kang2017object,lee2016multi} and another branch is feature-level aggregation \cite{stsn,Liu_2018_CVPR,wang2018manet,xiao2018matchtrans,zhu2018towards,zhu2017fgfa}. The former often applies a tracker to per-frame bounding box proposals over multiple frames to generate dense tubelets, making this category of approaches computationally expensive. Moreover, such methods are not trained end-to-end since the processes of per-frame proposal generation and bounding box tracking are independent. Feature aggregation improves detection by enhancing per-frame features through spatio-temporal aggregation of nearby frames. In this case, feature extraction and aggregation plus detection are trained in an end-to-end manner. We follow this elegant recipe and employ feature aggregation in our work. One natural way to execute feature aggregation for a reference frame is to estimate the motion across frames and warp the feature maps from neighboring frames to the reference one based on the motion. However, the results may suffer from robustness problem when the object appearances are deteriorated by motion blur or occlusion which often exists in videos as shown in Figure \ref{fig.fig1}(a). As such, we additionally introduce a stream of directly hallucinating/generating features through self-guided sampling from adjacent frames, which performs better than motion stream in this case. Nevertheless, the stream of self-guided sampling may fail to localize object accurately when object moves extremely fast (\emph{e.g.}, fast moving airplane in Figure \ref{fig.fig1}(b)). This is due to the fact that the receptive field in self-guided sampling for offset prediction is smaller than that in optical flow generation. Therefore, the range of estimated motion in sampling stream is shorter than that in motion stream, resulting in failure of motion capturing in sampling stream when object moves extremely fast. As a result, we simultaneously exploit both motion calibration and self-guided sampling in a two-stream feature aggregation structure for video object detection. More importantly, we novelly integrate feature aggregation into single shot object detection framework, which is more fit for computationally intensive video scenarios.

By consolidating the idea of two-stream feature aggregation into one-stage detection, we present a new Single Shot Video Object Detector (SSVD). Specifically, SSVD consists of three core modules: Feature Pyramid Network (FPN), two-stream feature aggregation structure and class/box subnets. FPN is taken as backbone network to output feature maps of multiple scales in a spatial pyramid. Each feature map in the pyramid is then fed into two-stream feature aggregation structure. One is motion stream, which estimates the displacements of objects across frames according to optical flow and warps the feature maps of nearby frames to the reference one along the motion paths. The other is sampling stream, which directly hallucinates the feature map of the reference frame by spatio-temporal sampling features from adjacent frames through deformable convolutions. The aggregated feature map in each stream is input into class/box subnets to classify anchor boxes and regress from anchor boxes to ground-truth object boxes. The final results are a blend of outputs on all the feature maps in the two streams. The whole SSVD is end-to-end trained by minimizing the focal loss for box classification plus the standard smooth $L_1$ loss for box regression.

The main contribution of this work is the proposal of a one-stage detector SSVD for addressing the issue of video object detection. Our SSVD, on one hand, takes advantages of single shot detectors (\emph{i.e.}, potentially faster and simpler than two-stage detectors), and on the other, leverages temporal coherence across frames to boost detection. The solution also leads to the elegant view of how temporal coherence along the motion paths and feature sampling across frames should be amended for feature aggregation, which are problems not yet fully understood in the literature.

\section{Related Work}

We briefly divide the most existing algorithms for object detection into two categories: object detection in images and object detection in videos.

\subsection{Object Detection in Images}
Inspired by the recent advances in image representation using deep Convolutional Neural Networks (CNN) \cite{he2016resnet,krizhevsky2012imagenet,Simonyan:ICLR15,Szegedy:CVPR15}, remarkable progresses have been witnessed for object detection \cite{dai2016r,hu2018relation,lin2017focal,liu2016ssd,ren2015faster,girshick2015fast,girshick2014rich,li2017light,redmon2016you,redmon2017yolo9000,chen2019adaptive,li2017scale,wang2018pedestrian}. In particular, one common deep solution for object detection is based on a two-stage paradigm, \emph{i.e.}, first perform region proposal and then do classification. R-CNN \cite{girshick2014rich} is one of first attempts that tackles object detection problem in a two-stage solution, which firstly utilizes selective search to generate region proposals and then classifies each proposal. Later on, SPP-Net \cite{he2015spatial} and Fast R-CNN \cite{girshick2015fast} extend \cite{girshick2014rich} by devising SPP pooling or ROI pooling to enable the sharing of features across region proposals, which significantly speed up the process of detection. Faster R-CNN \cite{ren2015faster} further advances Fast R-CNN by leveraging Region Proposal Networks (RPN) instead of selective search at the first stage. Compared to the costly per-region classification subnet in Faster R-CNN, R-FCN \cite{dai2016r} capitalizes on the fully convolutional network with position-sensitive ROI pooling. In addition, inspired by domain adaptation \cite{pan2019transferrable,pan2020open} for recognition, \cite{cai2019exploring,khodabandeh2019robust} focus on learning robust and domain-invariant detectors based on two-stage approaches.

Another direction mainly constructs one-stage detectors by omitting region proposal stage. One of the early successes is OverFeat \cite{sermanet2014overfeat} which constructs multi-scale sliding windows to jointly classify and localize objects in deep architecture. In \cite{redmon2016you}, YOLO divides feature maps into rigid grids, and objects are assigned to be detected by each grid. Technically, the prediction of objectness, confidence score of multiple classes and relative bounding boxes coordinates are devised as a regression problem in YOLO. As another genre of one-stage object detector, SSD \cite{liu2016ssd} further utilizes multiple feature maps at different scales with predefined default boxes to boost detection for objects in variety of scales and aspect ratios. Although methods with one-stage paradigm are potential to be faster and simpler, they trail in accuracy compared to two-stage ones. To match the performance of two-stage models, \cite{lin2017focal} presents an effective dense one-stage detector, RetinaNet, and equips it with Focal Loss to alleviate the foreground-background class imbalance and successfully achieve comparable performance of state-of-the-art two-stage detectors. More recently, anchor-free one stage detectors proposed in \cite{Kong_FoveaBox,Tian_FCOS} are superior in generalization ability since getting rid of limitation to the design of anchors.

\begin{figure*}
  \centering
  \includegraphics[width=0.9\linewidth]{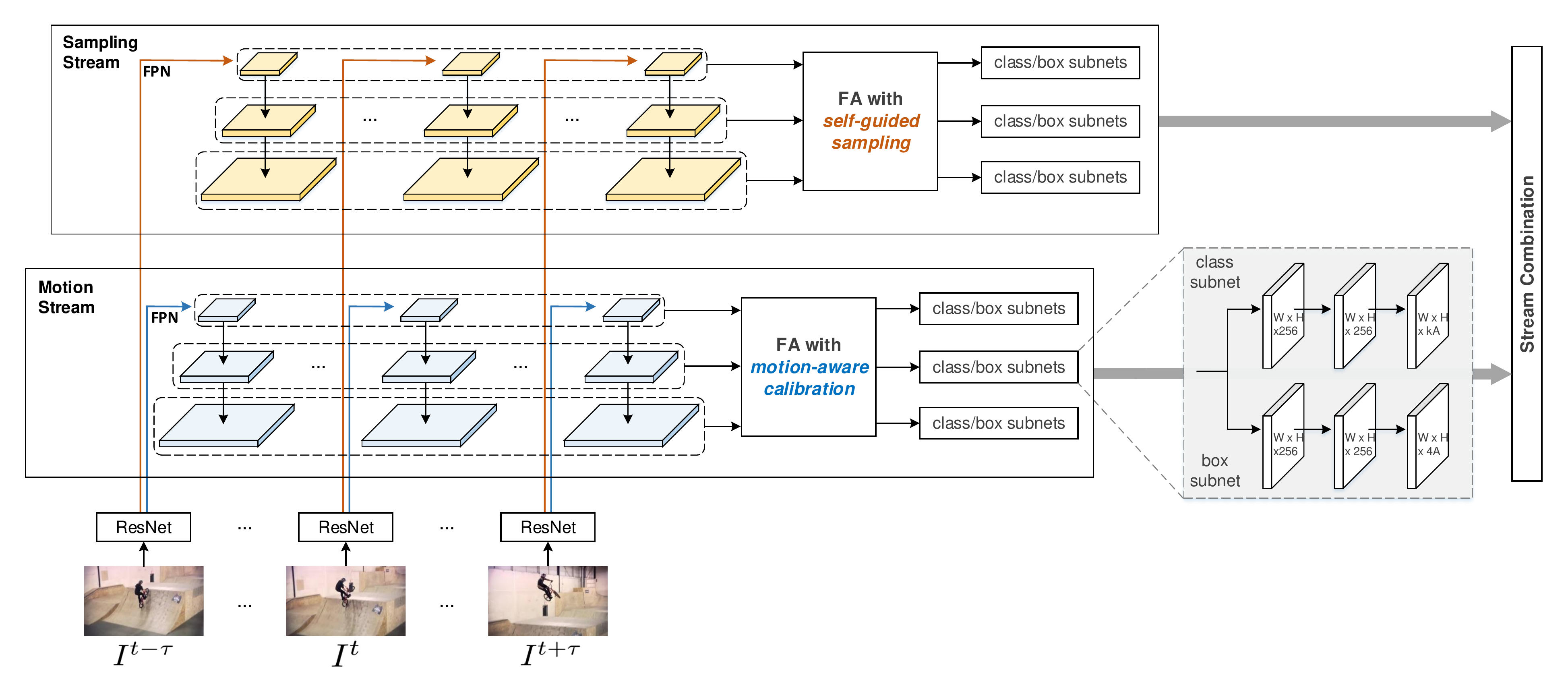}
  \vspace{-0.25in}
  \caption{An architecture overview of our Single Shot Video Object Detector (SSVD) (better viewed in color). It consists of three modules: (1) Feature Pyramid Network (FPN), (2) two-stream feature aggregation structure, and (3) Class/Box Subnets. Given the input sequence of adjacent frames $\{I^{t+\tau}\}^K_{\tau=-K}$, $K$ preceding frames and $K$ subsequent frames are taken as support frames for object detection in the reference frame $I^t$. Feature Pyramid Network is first leveraged to extract multi-scale pyramidal feature maps of each frame for motion and sampling stream separately. Next, feature aggregation (FA) with motion-aware calibration is employed along motion path by warping the feature maps of support frames in the same scale into the reference frame, whilst in the sampling stream we utilize feature aggregation (FA) with self-guided sampling to directly hallucinate the feature map of reference frame through spatio-temporal sampling features from support frames. After that, the aggregated feature map on each scale is injected into class/box subnets which include two parallel branches, one for anchor boxes classification and another for bounding box regression (k: the number of classes, A: the number of anchors per spatial location). The whole system integrates the three modules in one feed-forward CNN, making our detector single shot trainable. At inference, we blend all the predicted bounding boxes from two streams to produce the final detection results.}\label{fig.framework}
  \vspace{-0.15in}
\end{figure*}

\subsection{Object Detection in Videos}
Generalizing the existing detectors from still image to video domain is very challenge as video is an information-intensive media with both the spatial and temporal complex variations, not to mention that frames in videos are usually deteriorated by motion blur or occlusion.
The research on object detection in videos has proceeded along two different directions: box-level tracking \cite{feichtenhofer2017detect,han2016seq,kang2017object} and feature-level aggregation \cite{stsn,xiao2018matchtrans,zhu2017fgfa}. Box-level tracking employs box-level operations and post-processing over per-frame bounding box proposals to generate dense tubelets for identifying objects across frames. For instance, Seq-NMS \cite{han2016seq} builds a temporal graph across clips and then seeks the optimal path in this graph via dynamic programming for the selection of tubelets. Later, \cite{kang2017object,kang2017t,Kang_2016_CVPR} integrate per-frame proposals into tubelets for re-scoring, which further improves the robustness of video object detection. \cite{feichtenhofer2017detect} extends R-FCN with a tracking module for simultaneous detection and tracking. DorT \cite{luo2019detect} capitalizes on a scheduler network to switch between detection and tracking networks. Furthermore, HQ-link \cite{tang2019object} devises the cuboid proposal networks and Tubelet NMS to enable the global post-processing for robust video object detection.

Unlike box-level tracking that associates bounding boxes across frames with independent processes of linking/tracking, feature-level aggregation naturally enhances per-frame features via spatio-temporal aggregation, enabling an end-to-end detection paradigm. \cite{zhu2017fgfa,zhu2017dff} calibrate a sequence of per-frame feature maps with the guidance from optical flow and aggregate them along motion paths to enhance object detection. Next, \cite{wang2018manet} extends \cite{zhu2017fgfa} by exploiting additional motion path with box-level calibration for feature aggregation. Instead of estimating the motion across frames for warping the feature map, STSN \cite{stsn} performs object detection in a frame by learning to spatially sample features from adjacent frames for aggregation. STMN \cite{xiao2018matchtrans} adopts spatiotemporal memory module with spatial alignment mechanism to model long-term temporal appearance and motion dynamics. Besides, RDN \cite{deng2019relation} and SELSA \cite{wu2019sequence} strengthen region-level features by exploiting the relation/affinity between region proposals across frames

\begin{table}
\centering
  \small
  \setlength{\tabcolsep}{11.5pt}
  \caption{The main acronyms and notations used in this paper.}
  \begin{tabular}{c|l}
    \hline
    \hline
    $I^t$  & a frame at time t\\
    \hline
    $\{I^{t}\}$ & a set of frames \\
    \hline
    $K$ & temporal spanning range \\
    \hline
    \hline

    $\mathcal{N}_{\text{FPN}}^{\text{mo}}(\cdot)$ & feature pyramid network in motion stream \\
    \hline
    $\mathcal{N}_{\text{FPN}}^{\text{sp}}(\cdot)$ & feature pyramid network in sampling stream \\
    \hline

    $\mathcal{N}_{\text{pwc}}(\cdot)$ & PWC-Net \\
    \hline
    $\mathcal{N}_{\text{off}}(\cdot)$ & offset predictor \\
    \hline
    $\mathcal{N}_{\text{det}}(\cdot)$ & detect subnets \\
    \hline
    $\mathcal{W}(\cdot)$ & bilinear warping \\
    \hline
    $Dconv(\cdot)$ & deformable convolution layer \\
    \hline
    \hline

    $\{\textbf{\textit{f}}_{Pi}^t \}^6_{i=3}$ & pyramid features in motion stream \\
    \hline
    $\{\textbf{\textit{g}}_{Pi}^t \}^6_{i=3}$ & pyramid features in sampling stream \\
     \hline
      $\hat{\textbf{\textit{f}}}_{Pi}^{\text{mo}}$ & aggregated feature in motion stream \\
    \hline
    $\hat{\textbf{\textit{g}}}_{Pi}^{\text{sp}}$ & aggregated feature in sampling stream \\
    \hline
    $\mathbf{P}_t^{mo}$ & predictions in motion stream before NMS \\
    \hline
    $\mathbf{P}_t^{sp}$ & predictions in sampling stream before NMS \\
    \hline
    $\mathbf{D}_t$ & final detections \\
    \hline
    \hline
  \end{tabular}
  \label{tab.notation}
  \vspace{-0.15in}
\end{table}

\subsection{Summary}
Our approach belongs to feature-level aggregation methods. The novelty of our SSVD is on the exploitation of two-stream (motion and sampling stream) feature aggregation in one-stage detection paradigm, which is seldom explored. The motion stream performs feature aggregation by estimating motion from optical flow and warping the feature maps of nearby frames to the reference one along motion paths, whilst the sampling stream is trained to directly hallucinates the feature map of the reference frame through spatio-temporal sampling from adjacent frames.

\begin{figure}
  \centering
  \vspace{-0.10in}
  \includegraphics[width=1.05\linewidth]{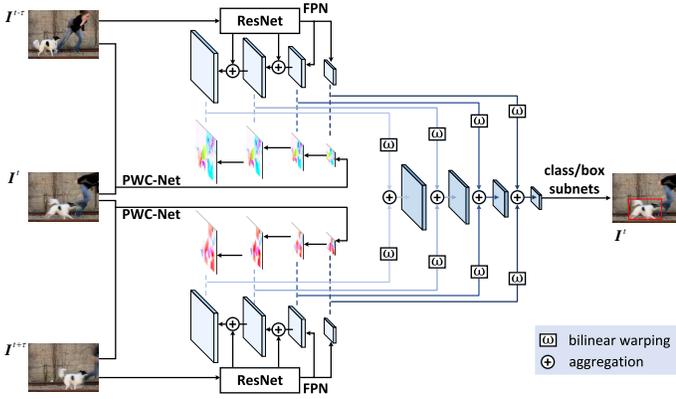}\\
  \vspace{-0.10in}
  \caption{Feature aggregation with motion-aware calibration in the motion stream of our SSVD. Specifically, each support frame is paired with reference frame and injected into PWC-Net to estimate optical flow in different resolution. The optical flow fields, in turn, guide the calibration of support feature maps out from Feature Pyramid Networks (FPN) by bilinear warping. Next, we directly average the calibrated feature maps of all support frames as aggregated feature map in each scale to boost object detection for reference frame. Note that here we only depict how to aggregate the support feature maps with motion-aware calibration, and omit the feature aggregation from reference frame for simplicity.} \label{fig.motion_fig}
  \vspace{-0.15in}
\end{figure}

\section{Single Shot Video Object Detector}
We devise our Single Shot Video Object Detector (SSVD) to facilitate object detection in videos by integrating two-stream feature aggregation via motion estimation and feature sampling into one-stage detection framework. An overview of SSVD is illustrated in Figure~\ref{fig.framework}. In particular, SSVD first utilizes Feature Pyramid Network to extract multi-scale pyramidal feature maps for motion and sampling stream separately. The pyramid feature maps of frames in the same scale are further aggregated by warping them into the reference one with motion-aware guidance along motion pathway or directly hallucinating the feature map of reference one with self-guided spatio-temporal sampling through sampling pathway, respectively. For the aggregated feature maps in each stream, the class/box subnets are leveraged to simultaneously classify objects and perform bounding box regression. As such, three core modules in our detector, \emph{i.e.}, Feature Pyramid Network, two-stream feature aggregation structure and class/box subnets are elegantly integrated into one feed-forward CNN, enabling single shot detection. In the inference stage, the final results are a blend of outputs on all the feature maps from two streams.

\subsection{Problem Formulation}
Suppose we have a sequence of adjacent frames $\{I^{t+\tau}\}^K_{\tau=-K}$, where the central frame $I^t$ is treated as the reference frame, and $K$ preceding frames plus $K$ subsequent frames are taken into account as the support frames. In the standard task of video object detection, the ultimate goal is to localize and recognize objects from the reference frame. Taking the inspiration from the temporal coherence exploration in video understanding \cite{li2018recurrent,li2018unified,long2019gaussian,pan2016learning,pan2016jointly,qiu2019learning,qiu2017learning,shi2020weakly}, we aim to detect the objects in the reference frame $I^t$ by additionally leveraging the spatio-temporal coherence distilled from the support frames. Technically, we formulate our video object detection architecture in one-stage paradigm with two-stream feature aggregation structure. The aggregated features are endowed with the characteristics of both temporal coherence distilled from the motion across adjacent frames in motion stream and contextual content encoded across the sampling features from adjacent frames in sampling stream.

Formally, given the reference frame $I^t$ and each support frame $I^{t+\tau}$, Feature Pyramid Network (FPN) \cite{lin2017feature} is leveraged to extract multi-scale pyramidal feature maps of each frame for motion and sampling stream, separately. More precisely, by feeding each frame into the FPN (\emph{i.e.}, $\mathcal{N}_{\text{FPN}}^{\text{mo}}$, $\mathcal{N}_{\text{FPN}}^{\text{sp}}$) in motion and sampling stream, we can obtain two sets of multi-scale pyramidal feature maps, \emph{i.e.}, $\{\textbf{\textit{f}}_{Pi}^t \}^6_{i=3}= \mathcal{N}_{\text{FPN}}^{\text{mo}} (I^t)$ and $\{\textbf{\textit{g}}_{Pi}^t \}^6_{i=3}= \mathcal{N}_{\text{FPN}}^{\text{sp}} (I^t)$\footnote{$P3,P4,P5$ denote the output of standard FPN, and correspond to the last residual blocks in conv3, conv4, conv5 of ResNet. $P6$ is produced by attaching $3\times3$ conv with stride 2 over the last residual block in conv5.}, for motion and sampling stream, respectively. The network structure of $\mathcal{N}_{\text{FPN}}^{\text{mo}}$ and $\mathcal{N}_{\text{FPN}}^{\text{sp}}$ in the two streams is identical but the parameters of the two FPN are not shared during training. These pyramidal feature maps decrease in scale progressively and allow predictions of detection at multiple scales. Next, motion/sampling stream takes the feature maps of support and reference frames at the same scale (\emph{i.e.}, $\{\textbf{\textit{f}}_{Pi}^{t+\tau},\textbf{\textit{f}}_{Pi}^t\}/\{\textbf{\textit{g}}_{Pi}^{t+\tau},\textbf{\textit{g}}_{Pi}^t\}$) as the inputs, and performs feature aggregation with motion-aware calibration or self-guided sampling. On the basis of the aggregated feature map of each scale in each stream, class/box subnets are utilized to simultaneously classify anchor boxes and regress from anchor boxes to ground-truth object boxes. The final results at inference are produced by combing the outputs from all scales in the two streams with late fusion scheme. The main acronyms and notations in our paper are given in Table \ref{tab.notation}.

\subsection{Motion Stream}\label{sub_sec2}
One natural way to enhance per-frame feature is to estimate the motion across frames in the form of optical flow and warp the features from neighboring frames to the target one based on the motion. Such way of motion compensation has convincingly demonstrated high capability of modeling temporal correlation in video super-resolution \cite{caballero2017real}, video object detection \cite{wang2018manet,zhu2017fgfa}, and video translation \cite{chen2019mocycle}. As a result, we follow this elegant recipe and design the module of feature aggregation with motion-aware calibration which performs feature aggregation over the input feature maps with same scale along motion stream, as illustrated in Figure \ref{fig.motion_fig}. The idea behind this feature aggregation module is to calibrate the feature map of support frame to the reference frame with the guidance from optical flow across them in the context of one-stage video object~detection.

\textbf{Motion Estimation.} Technically, with the input reference and support frames, we first estimate the motion between them in the form of optical flow. Unlike the works \cite{wang2018manet,zhu2017fgfa,zhu2017dff} which capitalize on FlowNet-s \cite{dosovitskiy2015flownet} to produce optical flow, PWC-Net \cite{sun2018pwc} is particularly remould in our motion stream. Compared to a generic U-Net CNN built in FlowNet-s, the pyramid processing in PWC-Net is more flexible to the calibration of multi-scale feature maps, tailored to one-stage detection paradigm. Furthermore, the architecture of PWC-Net is smaller in size and easier to be trained when integrating into object detector.

In particular, by injecting the reference frame $I^t$ and the support frame $I^{t+\tau}$ into PWC-Net (\emph{i.e.}, $ \mathcal{N}_{\text{pwc}}$), a set of multi-scale flow fields describing the motion between them are obtained: $\{\textbf{\textit{m}}_{Pi}^{(t, t+\tau)} \}^6_{i=3}= \mathcal{N}_{\text{pwc}} (I^t,I^{t+\tau})$. $\{\textbf{\textit{m}}_{Pi}^{(t, t+\tau)} \}^6_{i=3}$ denote the outputs from the last four groups of layers in PWC-Net by removing the refinement module and correspond to the support frame's feature map $\{\textbf{\textit{f}}_{Pi}^{t+\tau} \}^6_{i=3}$ in each scale, respectively. In practice, to align the scale of each flow field $\textbf{\textit{m}}_{Pi}^{(t, t+\tau)}$ with the corresponding feature map $\textbf{\textit{f}}_{Pi}^{t+\tau}$, we downsample the spatial resolution of the input frames in PWC-Net by a factor 2.

\textbf{Motion-aware Calibration.} For each scale, given the feature map of support frame $\textbf{\textit{f}}_{Pi}^{t+\tau}$ and the corresponding flow field $\textbf{\textit{m}}_{Pi}^{(t, t+\tau)}$, we conduct the motion-aware calibration by warping the feature map of support frame to the reference frame with the guidance of the flow:
\begin{equation}\label{feat_warping}
  \textbf{\textit{f}}_{Pi}^{t+\tau\rightarrow t} = \mathcal{W}(\textbf{\textit{f}}_{Pi}^{t+\tau}, \textbf{\textit{m}}_{Pi}^{(t, t+\tau)}),~~~i \in \{3, 4, 5, 6\},
  \end{equation}
where $\textbf{\textit{f}}_{Pi}^{t+\tau\rightarrow t}$ is the calibrated feature map of support frame $I^{t+\tau}$ and $\mathcal{W}(\cdot)$ denotes the function of bilinear warping.

\textbf{Feature Aggregation.} After obtaining the calibrated feature maps of all support frames, we directly average them as the aggregated feature map in each scale of the reference frame in motion stream:
\begin{equation}\label{aggre_flow}
  \hat{\textbf{\textit{f}}}_{Pi}^{\text{mo}} = \frac{\sum_{\tau=-K}^K \textbf{\textit{f}}^{t+\tau\rightarrow t}_{Pi}}{2K+1},~~~~~i \in \{3, 4, 5, 6\}.
\end{equation}
Through feature aggregation with motion-aware calibration, the aggregated feature $\hat{\textbf{\textit{f}}}_{Pi}^{\text{mo}}$ of reference frame is enhanced with the temporal coherence distilled in motion.

\subsection{Sampling Stream}\label{sub_sec3}
\begin{figure}
  \centering
  \includegraphics[width=0.95\linewidth]{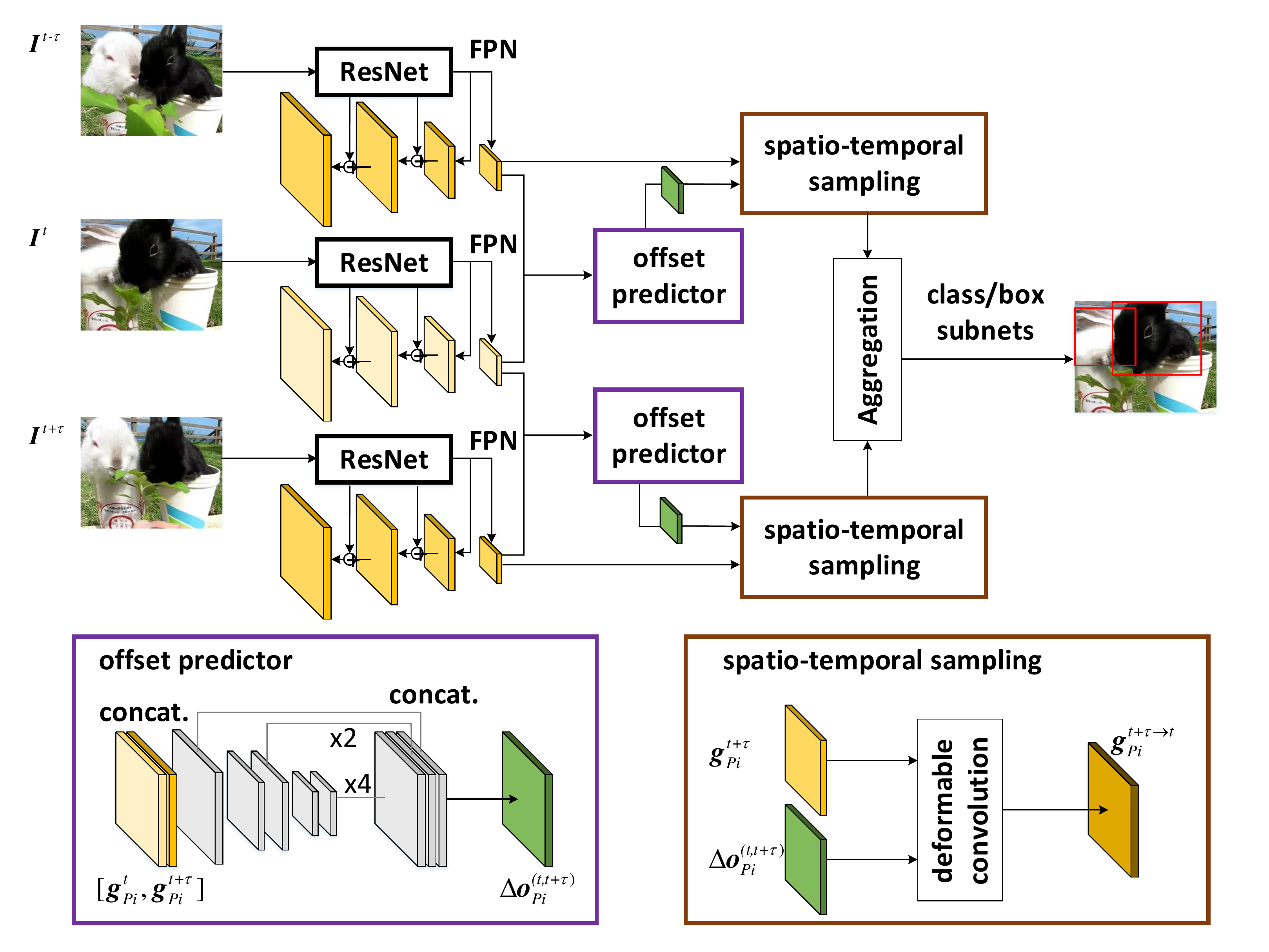}\\
  \vspace{-0.10in}
  \caption{Feature aggregation with self-guided sampling in the sampling stream of our SSVD (`concat': `channel-wise concatenate'). Note that here we only take the feature map with the lowest resolution as an example to depict the detailed processing of self-guided sampling. Specifically, for each spatial resolution, we concatenate the feature map from each support frame with feature map of reference frame, which will be fed into offset predictor for predict the offsets between the reference and support frames. Next, spatio-temporal sampling is further performed to directly hallucinate the feature of reference frame via deformable convolution over the feature map of support frame and the predicted offsets.} \label{fig.sample_fig}
  \vspace{-0.30in}
\end{figure}

In the motion stream, we facilitate the feature aggregation by exploring the explicit motion compensation fully conditioned on optical flow across adjacent frames. Nevertheless, such design in motion path is heavily restricted by the quality of estimated motion and may suffer from robustness problem when the object appearances are deteriorated by motion blur or occlusion. To alleviate this issue, we devise feature aggregation module with self-guided sampling in sampling stream that directly hallucinates features via spatio-temporal sampling from the support frames, as shown in Figure \ref{fig.sample_fig}. The philosophy is originated from the idea of deformable convolution \cite{dai2017deformable} which performs non-rigid spatial sampling with self-learnt offsets. More precisely, deformable convolution upgrades standard convolution by adding 2D offsets to the regular grid sampling locations, where the 2D offsets are inferred by the input feature itself, without additional supervision. In our case, we extend the augmentation of spatial sampling locations in standard deformation convolution \cite{dai2017deformable} which is only conditioned on one feature map to the measure of deformation across two feature maps of the reference frame and the support frame. In other words, our feature aggregation module in sampling stream learns to predict the implicit correlation between support and reference frames in the form of 2D offsets depending on the input frames, without referring to the estimated optical flow. As such, the hallucinated feature of reference frame for aggregation is acquired by spatio-temporal sampling over support frames through deformable convolutions with the guidance of the self-learnt 2D offset.

\textbf{Offset Predictor.} To learn the deformation between reference and support frames, we uniquely present a new architecture of our offset predictor in feature aggregation module with self-guided sampling to predict the offsets. Here the predicted offsets implicitly depict the temporal correlation between the reference and support frames, which could further guide the following spatio-temporal sampling from support frames. In the architecture of our offset predictor, we adopt the ``U-Net" structure \cite{ronneberger2015u} design to process input feature maps across multiple scales and enrich the receptive field for offset prediction. Compared to \cite{stsn} that adopts four stacked deformable convolution module for offset prediction, the design of our offset predictor contains $\sim$40 times less parameters, which is more efficient and tailored to one-stage video object detector.

Concretely, given the feature maps in each scale of reference frame $\textbf{\textit{g}}_{Pi}^{t}$ and support frame $\textbf{\textit{g}}_{Pi}^{t+\tau}$ in sampling stream, we firstly concatenate the two feature maps along channel dimension, which is taken as the input of offset predictor ($\mathcal{N}_{\text{off}}$). Next, the concatenated feature map is fed into three groups of convolutional layers, where the first group retains the scale of the concatenated feature map and the last two groups progressively downsample the scale by a factor 2. Finally, we concatenate the output feature maps of three convolution groups in offset predictor with upsampling operation, and leverage them to predict the offsets $\Delta\textbf{\textit{o}}_{Pi}^{(t, t+\tau)} = \mathcal{N}_{\text{off}}(\textbf{\textit{g}}_{Pi}^{t+\tau}, \textbf{\textit{g}}_{Pi}^{t})$ for $\textbf{\textit{g}}_{Pi}^{t+\tau}$ at each scale.

\textbf{Spatio-temporal Sampling.} Given the feature map of support frame $\textbf{\textit{g}}_{Pi}^{t+\tau}$ and its predicted offset $\Delta\textbf{\textit{o}}_{Pi}^{(t, t+\tau)}$, we hallucinate the feature of reference frame $\textbf{\textit{g}}_{Pi}^{t+\tau\rightarrow{t}}$ via deformable convolution over $\textbf{\textit{g}}_{Pi}^{t+\tau}$. Here, we directly exploit the predicted offset from our offset predictor as the 2D offset in deformable convolution. Such way enables the spatio-temporal sampling from support frames with respect to the reference frame and the generated feature is therefore tailored to the spatio-temporal context.

\textbf{Feature Aggregation.} After obtaining the hallucinated feature maps of reference frame from all support frames in each scale, we average them as the aggregated feature map of reference frame in sampling stream:
\begin{equation}\label{aggre_sp}
  \hat{\textbf{\textit{g}}}_{Pi}^{\text{sp}} = \frac{\sum_{\tau=-K}^K \textbf{\textit{g}}^{t+\tau\rightarrow{t}}_{Pi}}{2K+1},~~~~~i \in \{3, 4, 5, 6\}.
\end{equation}
Through feature aggregation with self-guided sampling, the aggregated feature $\hat{\textbf{\textit{g}}}_{Pi}^{\text{sp}}$ is endowed with the contextual content encoded in sampling.

\begin{figure}
  \centering
  \vspace{-0.1in}
  \includegraphics[width=0.88\linewidth]{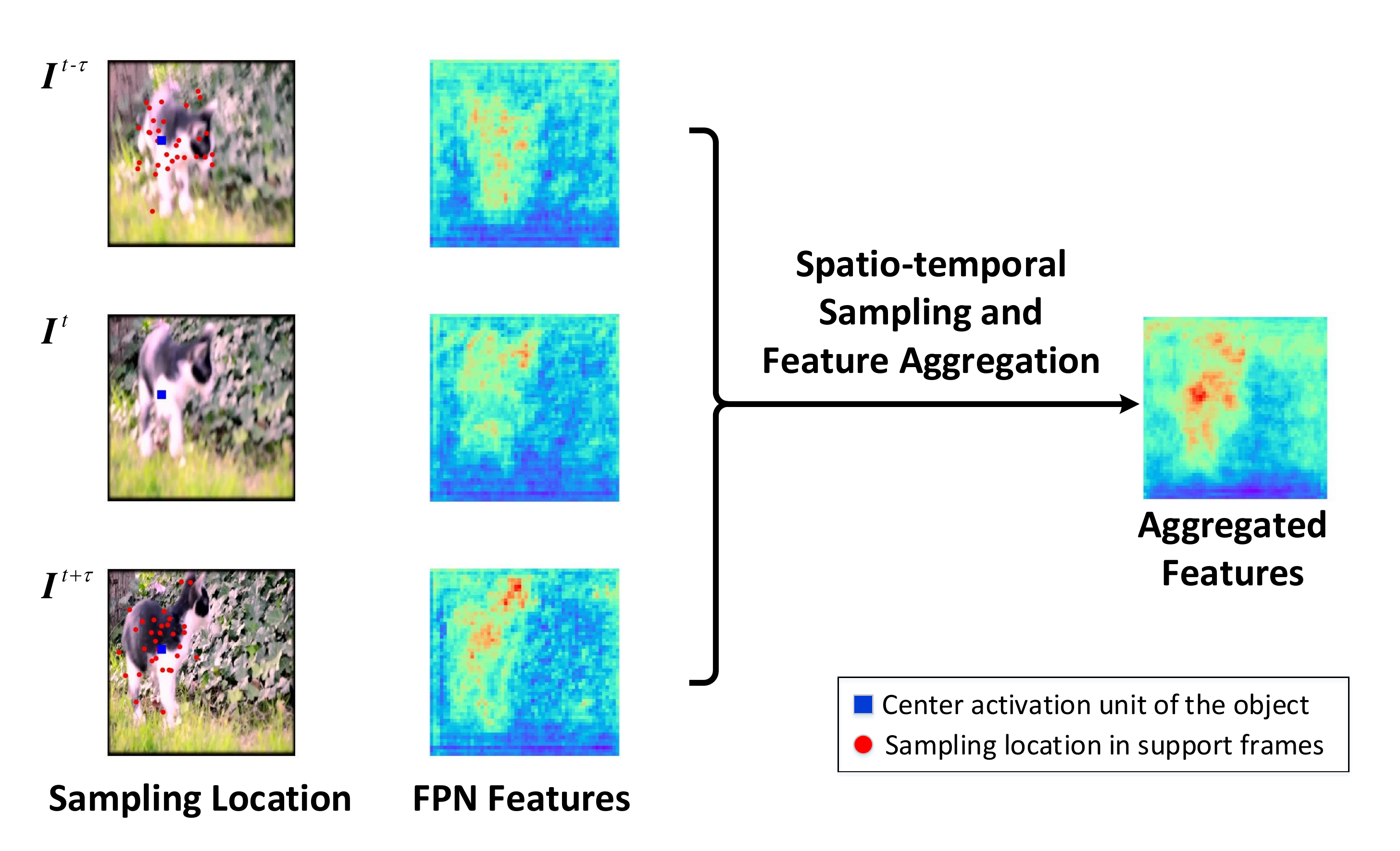}\\
  \vspace{-0.15in}
  \caption{Intermedia data of our Sampling Stream (better viewed in color). The blue square in reference frame depicts a pixel, for which we want to compute a convolution output. The red points are the corresponding sampling locations in support frames, which are predicted by our offset predictor. } \label{fig.intermedia}
  \vspace{-0.20in}
\end{figure}

\textbf{Intermedia Data.} In Figure \ref{fig.intermedia}, we provide some intermedia data to illustrate the effect of feature aggregation with self-guided sampling. The blue square in reference frame depicts a pixel, for which we want to compute a convolution output. The red points are the corresponding sampling locations in support frames, which are predicted by our offset predictor. As illustrated in this figure, our offset predictor learns to sample the locations from the supporting frames that belong to the corresponding object. The aggregated feature of reference frame is thus hallucinated by performing deformable convolution over the FPN feature map of support frame and the predicted offsets. From the aggregated feature map of reference frame, we can clearly observe that the features depicting the same object are strengthened.

\subsection{Training and Inference \label{sub_sec4}}

During training, the aggregated feature in each scale (\emph{i.e.}, $\{\hat{\textbf{\textit{f}}}_{Pi}^{\text{mo}}\}^6_{i=3}$ and $\{\hat{\textbf{\textit{g}}}_{Pi}^{\text{sp}}\}^6_{i=3}$) in both motion and sampling stream is injected into class/box subnets, which simultaneously classify anchor boxes with Focal Loss \cite{lin2017focal} ($\mathcal{L}_{\text{FL}}$) and regress from anchor boxes to ground-truth object boxes with Smooth $L_1$ Loss \cite{girshick2015fast} ($\mathcal{L}_{\text{Loc}}$). Accordingly, the overall objective of our SSVD is computed as
\begin{equation}\label{equ.loss_all}\small
  \begin{split}
  \mathcal{L} = \frac{1}{N_{\text{fg}}}\{\sum_{j} \sum_{c} \mathcal{L}_{\text{FL}}(p_{j,c}^{\text{mo}}, l_j^{\ast})+\sum_{j} \sum_{c} \mathcal{L}_{\text{FL}}(p_{j,c}^{\text{sp}}, l_j^{\ast})\\
  +[l_j^{\ast}\ge{1}]\sum_{j} \mathcal{L}_{\text{Loc}}(x_i^{\text{mo}}, t_j^{\ast})+[l_j^{\ast}\ge{1}]\sum_{j} \mathcal{L}_{\text{Loc}}(x_j^{\text{sp}}, t_j^{\ast})\},\\
  \end{split}
\end{equation}
where $j$ is the index of anchor in a mini-batch, $l_j^{\ast}$ is the ground truth class label and $[l_j^{\ast}\ge{1}]$ indicates that the $j$-th anchor is assigned to one of the foreground classes. $t_j^{\ast}$ denotes the location and size of ground-truth object boxes. $N_\text{{fg}}$ is the number of anchors assigned to ground-truth object boxes. $p_{j,c}^{\text{mo}}$ and $p_{j,c}^{\text{sp}}$ denotes the predicted confidence score of class $c$ of the $j$-th anchor in motion and sampling stream, respectively. $x_j^{\text{mo}}$ and $x_j^{\text{sp}}$ is the predicted coordinates of the $j$-th anchor in motion and sampling stream. Note that not all the frames within aggregation range are taken in the training phase, we just randomly select two of them. The detailed training process is presented in Algorithm \ref{alg.ssvd_train}.

\begin{algorithm}[t]
  \caption{Training Process of our SSVD}
  \small
  \begin{algorithmic}[1] 
  \State \textbf{Input}:  a reference frame  $I^t$ and two support frames $I^{t+\tau_1}$, $I^{t+\tau_2}$ sampled from frame $\{I^{t+\tau}\}_ {\tau=-K}^{K}$.
  \State \textbf{Pre-processing}: Data augmentation is performed as in \cite{liu2016ssd}.
  \State \textbf{Feature Extraction}:
  \State \quad \textbf{for} $\tau \in \{0,\tau_1, \tau_2\}$ \textbf{do}
  \State \qquad $\{\textbf{\textit{f}}_{Pi}^{t+\tau} \}^6_{i=3}= \mathcal{N}_{\text{FPN}}^{\text{mo}} (I^{t+\tau})$ \Comment{Motion Stream}
  \State \qquad $\{\textbf{\textit{g}}_{Pi}^{t+\tau} \}^6_{i=3}= \mathcal{N}_{\text{FPN}}^{\text{sp}} (I^{t+\tau})$ \Comment{Sampling Stream}
  \State \quad \textbf{end for}
  \State \textbf{FA with Motion-aware Calibration}:
     \State \quad \textbf{for} $\tau \in \{\tau_1, \tau_2\}$ \textbf{do}
     \State \qquad $\{\textbf{\textit{m}}_{Pi}^{(t, t+\tau)} \}^6_{i=3}= \mathcal{N}_{\text{pwc}} (I^t,I^{t+\tau})$ \Comment{Motion Estimation}
     \State \qquad $\textbf{\textit{f}}_{Pi}^{t+\tau\rightarrow t} = \mathcal{W}(\textbf{\textit{f}}_{Pi}^{t+\tau}, \textbf{\textit{m}}_{Pi}^{(t, t+\tau)}),i\in\{3,4,5,6\}$ \Comment{Calibration}
     \State \quad \textbf{end for}
     \State \quad $\hat{\textbf{\textit{f}}}_{Pi}^{\text{mo}} = \frac{ \textbf{\textit{f}}^{t+\tau_1\rightarrow t}_{Pi}+\textbf{\textit{f}}^{t+\tau_2\rightarrow t}_{Pi}}{2},i \in \{3, 4, 5, 6\}$ \Comment{Aggregation}
  \State \textbf{FA with Self-guided Sampling}:
     \State \quad \textbf{for} $\tau \in \{\tau_1, \tau_2\}$ \textbf{do}
     \State \qquad \textbf{for} $i \in \{3,4,5,6\}$ \textbf{do}
     \State \quad \qquad $\Delta\textbf{\textit{o}}_{Pi}^{(t, t+\tau)} = \mathcal{N}_{\text{off}}(\textbf{\textit{g}}_{Pi}^{t+\tau}, \textbf{\textit{g}}_{Pi}^{t})$ \Comment{Offset Prediction}
     \State \quad \qquad $\textbf{\textit{g}}_{Pi}^{t+\tau\rightarrow t} = DConv(\textbf{\textit{g}}_{Pi}^{t+\tau}, \textbf{\textit{o}}_{Pi}^{(t, t+\tau)})$ \Comment{Sampling}
     \State \qquad \textbf{end for}
     \State \quad \textbf{end for}
     \State \quad $\hat{\textbf{\textit{g}}}_{Pi}^{\text{sp}} = \frac{\textbf{\textit{g}}^{t+\tau_1\rightarrow{t}}_{Pi}+\textbf{\textit{g}}^{t+\tau_2\rightarrow{t}}_{Pi}}{2},i \in \{3, 4, 5, 6\}$ \Comment{Aggregation}
  \State \textbf{Perform Detection}:
  \State \quad Applying detect subnets on $\hat{\textbf{\textit{f}}}_{Pi}^{\text{mo}}$ and $\hat{\textbf{\textit{g}}}_{Pi}^{\text{sp}}$.
  \State \quad Calculate objectives as in Equation \ref{equ.loss_all}.
  \State \textbf{Optimizing with stochastic gradient descent}.
  \end{algorithmic}
  \label{alg.ssvd_train}
  \end{algorithm}

At the inference stage, we adopt the late fusion scheme to combine the detection results from motion and sampling streams. Specifically, we first accumulate all the predicted bounding boxes from motion and sampling streams and then apply Non-Maximum Suppression (NMS) with a threshold of jaccard overlap 0.45 per class to produce the final results. We detail the inference algorithm in  Algorithm \ref{alg.ssvd_inference}. Moreover, implementation details of training and inference are given in the experimental section.

\section{Experiments}

\subsection{Dataset Sampling and Evaluation}

We evaluate our SSVD on the large-scale benchmark for video object detection task, \emph{i.e.}, ImageNet \cite{ILSVRC15} object detection from video (VID) dataset, which contains 3,862 training and 555 validation videos in 30 classes. As the annotations of testing videos are not publicly available, we follow the widely adopted protocols in \cite{feichtenhofer2017detect,Kang_2016_CVPR,zhu2017fgfa} to report the results on validation set in terms of the evaluation metric of mean Average Precision (mAP). The 30 object classes in ImageNet VID dataset are a subset of 200 classes of ImageNet object detection (DET) dataset. Therefore, we follow \cite{feichtenhofer2017detect,zhu2017fgfa} and train SSVD on the intersection of ImageNet VID and ImageNet DET. Due to the redundancy among frames in ImageNet VID and the large variations in the number of samples per class in ImageNet DET, we sample 15 frames from each video in ImageNet VID dataset and at most 2,000 images per class in ImageNet DET dataset for~training.

\subsection{Model Architecture Design}
\textbf{Feature Pyramid Network.} FPN is built at the top of ResNet-101 pre-trained on ImageNet. As in \cite{lin2017feature}, P3, P4, and P5 are computed from the outputs of ResNet residual stage (conv3 to conv5) with top-down pathway and lateral connections. P6 is obtained by attaching a $3\times3$ stride-2 convolution over the last residual block in conv5. As such, with the input size of $448^2$ in SSVD, the spatial scale of P3, P4, P5, and P6 is $56^2, 28^2, 14^2,$ and $7^2$, respectively.

\textbf{Two-stream Feature Aggregation.} For motion stream, we utilize PWC-Net \cite{sun2018pwc} pre-trained on Flying Chairs dataset for optical flow estimation. For sampling stream, each conv layer in offset predictor consists of $3\times3$ kernel with 256 filters except for the final one with 72 filters. All the parameters of offset predictor are randomly initialized. The deformable group in deformable conv is set as 4.

\begin{algorithm}[t]
  \caption{Inference Algorithm of our SSVD}
  \small
  \begin{algorithmic}[1] 
  \State \textbf{Input}: video frames $\{I^t\}$, temporal spanning range $K$.
  \State \textbf{Feature Buffer Initialization}:
  \State \ \ \textbf{for} {$t=1$ \textbf{to} $K+1$} \textbf{do}
  \State \ \ \ \  $\{\textbf{\textit{f}}_{Pi}^{t+\tau} \}^6_{i=3}= \mathcal{N}_{\text{FPN}}^{\text{mo}} (I^{t+\tau})$ \Comment{Motion Stream}
  \State \ \ \ \  $\{\textbf{\textit{g}}_{Pi}^{t+\tau} \}^6_{i=3}= \mathcal{N}_{\text{FPN}}^{\text{sp}} (I^{t+\tau})$ \Comment{Sampling Stream}
  \State \ \  \textbf{end for}
  \State \textbf{Online Detection}:
  \State \ \ \textbf{for} {$t=1$ \textbf{to} $\infty$} \textbf{do}

   \State \ \ \ \ \textbf{(In Motion Stream)}
    \State \ \ \ \ \textbf{for} $\tau=max(1, t-K)$ \textbf{to} $t+K$ {do}
      \State \ \ \ \ \ \ $\{\textbf{\textit{m}}_{Pi}^{(t, t+\tau)} \}^6_{i=3}= \mathcal{N}_{\text{pwc}} (I^t,I^{t+\tau})$  \Comment{Motion Estimation}
      \State \ \ \ \ \ \ $\textbf{\textit{f}}_{Pi}^{t+\tau\rightarrow t} = \mathcal{W}(\textbf{\textit{f}}_{Pi}^{t+\tau}, \textbf{\textit{m}}_{Pi}^{(t, t+\tau)}),i\in\{3,4,5,6\}$ \Comment{Calibration}
    \State \ \ \ \ \textbf{end for}
    \State \ \ \ \ $\hat{\textbf{\textit{f}}}_{Pi}^{\text{mo}} = \frac{\sum_{\tau=-K}^K \textbf{\textit{f}}^{t+\tau\rightarrow t}_{Pi}}{2K+1},~i \in \{3, 4, 5, 6\}$ \Comment{Aggregation}
    \State \ \ \ \ $\mathbf{P}_t^{mo}=\mathcal{N}_{det}^{mo}(\hat{\textbf{\textit{f}}}_{Pi}^{\text{mo}})$ \Comment{Classification and Regression}

    \State \ \ \ \ \textbf{(In Sampling Stream)}
    \State \ \ \ \ \textbf{for} $\tau=max(1, t-K)$ \textbf{to} $t+K$ {do}
    \State \ \ \ \ \ \ \textbf{for} $i\in\{3,4,5,6\}$ \textbf{do}
    \State \ \ \ \ \ \ \ \  $\Delta\textbf{\textit{o}}_{Pi}^{(t, t+\tau)} = \mathcal{N}_{\text{off}}(\textbf{\textit{g}}_{Pi}^{t+\tau}, \textbf{\textit{g}}_{Pi}^{t})$ \Comment{Offset Prediction}
    \State \ \ \ \ \ \ \ \  $\textbf{\textit{g}}_{Pi}^{t+\tau\rightarrow t} = DConv(\textbf{\textit{g}}_{Pi}^{t+\tau}, \textbf{\textit{o}}_{Pi}^{(t, t+\tau)})$ \Comment{Sampling}
   \State \ \ \ \ \ \ \textbf{end for}
    \State \ \ \ \ \textbf{end for}
    \State \ \ \ \ $\hat{\textbf{\textit{g}}}_{Pi}^{\text{sp}} = \frac{\sum_{\tau=-K}^K \textbf{\textit{g}}^{t+\tau\rightarrow{t}}_{Pi}}{2K+1}, i \in \{3, 4, 5, 6\}$ \Comment{Aggregation}
    \State \ \ \ \ $\mathbf{P}_t^{sp}=\mathcal{N}_{det}^{sp}(\hat{\textbf{\textit{g}}}_{Pi}^{\text{sp}})$ \Comment{Classification and Regression}
    \State \ \ \ \ \textbf{Merge $\mathbf{P}_t^{mo}$, $\mathbf{P}_t^{sp}$ together as $\mathbf{P}_t$.}
    \State \ \ \ \ $\mathbf{D}_t=\text{NMS}(\mathbf{P}_t)$    \Comment{Performing NMS}

   \State \ \ \ \  $\{\textbf{\textit{f}}_{Pi}^{t+K+1} \}^6_{i=3}= \mathcal{N}_{\text{FPN}}^{\text{mo}} (I^{t+K+1})$ \Comment{Motion Stream}
   \State \ \ \ \  $\{\textbf{\textit{g}}_{Pi}^{t+K+1}\}^6_{i=3}= \mathcal{N}_{\text{FPN}}^{\text{sp}} (I^{t+K+1})$ \Comment{Sampling Stream}
   \State \ \ \ \ \textbf{Update feature buffers.}

   \State \ \ \textbf{end for}
\State \textbf{Offline Post-processing (optional)}:
\State \ \ $\{\mathbf{D}_t\}=\text{Seq-NMS}(\{\mathbf{P}_t\})$ \Comment{Applying Seq-NMS \cite{han2016seq}}

  \end{algorithmic}
  \label{alg.ssvd_inference}
  \end{algorithm}

\textbf{Class/Box Subnets.} The class/box subnets include two parallel branches, \emph{i.e.}, class and box branch. The structure in class branch starts from two $3\times3$ conv layers, each with 256 filters, followed by a $3\times3$ conv layer with $kA$ filters plus sigmoid activations. Here $k$ is the number of classes and $A$ is the number of anchors per spatial location. The structure of box branch is identical to that in class branch except that box branch terminates in $4A$ linear outputs per spatial location. For each anchor, the four outputs denote the predicted relative offsets between the anchor and the ground-truth box. To handle different scales and aspect ratios of objects, translation-invariant anchors are assigned to P3, P4, P5, P6 with anchor areas ranging from $32^2$ to $256^2$. As in \cite{lin2017feature,lin2017focal}, each pyramid layer is associated with anchors at three aspect ratios of \{1:2, 1:1, 2:1\} and three size factors of \{$2^0, 2^{1/3}, 2^{2/3}$\}. As such, there are $A=9$ anchors per spatial location in total.

\subsection{Implementation Details}
\textbf{Training.} In our experiments, the whole architecture of our SSVD is trained over 4 GPUs by synchronized SGD with momentum of 0.9 and weight decay of $0.0001$. The batch size is set as 16. Following \cite{wang2018manet,zhu2017fgfa}, the two-phase training strategy is adopted. In the first phase, we construct a still image detector by directly linking the class/box subnets with Feature Pyramid Network, and train this detector over the images/frames from the combined training set of ImageNet DET and VID. The learning rate is set as $0.001$ in the first $80,000$ iterations and $0.0001$ in the last $40,000$ iterations. In the second phase, the whole SSVD is trained on ImageNet VID, with the learning rate of $0.001$ and $0.0001$ in the first $60,000$ iterations and the last $30,000$ iterations, respectively. Note that Feature Pyramid Network and class/box subnets are initialized from the weights learnt in the first phase. To make the model more robust and adaptive to variant changes in videos, we perform the widely-adopted data augmentation as in SSD \cite{liu2016ssd}. The aggregation range $K$ is set as 12 in all experiments. As in FGFA \cite{zhu2017fgfa}, we utilize temporal dropout in training stage by randomly selecting two support frames from the input adjacent frames $\{I^{t+\tau}\}^K_{\tau=-K}$ of reference frame $I^t$.

\textbf{Inference.} At inference, we follow \cite{zhu2017fgfa} and sequentially process each frame with a sliding feature buffer of the nearby frames. The capacity of feature buffer is maintained as 25 except for the beginning and ending 12 frames. For each feature buffer consisting of 24 support frames and one reference frame, we uniformly select 6 support frames and aggregate them to the reference frame for object detection.  Note that no data augmentation is adopted during inference. For offline post-processing, we leverage a widely-adopted global linking and suppression algorithm (\emph{i.e.} Seq-NMS \cite{han2016seq}) to perform box-level association.

\begin{table}[t]
  \centering
  \vspace{-0.1in}
  \footnotesize
  \setlength{\tabcolsep}{3.2pt}
  \caption{\small Performance contribution of each stream in our SSVD on ImageNet VID validation set. The subscripts denote the absolute performance gains compared to the single-frame baseline~(a).}
  \vspace{-0.1in}
  \begin{tabular}{ccccccc}
    \hline
      & \multicolumn{2}{c|}{\textbf{feature aggregation}} &\multicolumn{4}{c}{\textbf{~~~~mean average precision (\%)~~~~}}\\
    \hline
     &\multicolumn{1}{c}{motion} & \multicolumn{1}{c|}{sampling} & \multirow{2}*{slow~~~~~~} & \multirow{2}*{medium~~} & \multirow{2}*{fast~~~~~~~} & \multirow{2}*{overall~~}\\

     &\multicolumn{1}{c}{stream} & \multicolumn{1}{c|}{stream} &  &  & & \\
     \hline
     \multicolumn{1}{c}{(a)} & \multicolumn{1}{c}{} & \multicolumn{1}{c|}{} & \multicolumn{1}{l}{$81.1$} & \multicolumn{1}{l}{$73.7$} & \multicolumn{1}{l}{$52.8$} & \multicolumn{1}{l}{$74.5$} \\
     \hline
     \multicolumn{1}{c}{(b)} & \multicolumn{1}{c}{\checkmark} & \multicolumn{1}{c|}{} & \multicolumn{1}{l}{$85.4_{\uparrow 4.3}$}  & \multicolumn{1}{l}{$77.5_{\uparrow 3.8}$} &  \multicolumn{1}{l}{$56.1_{\uparrow 3.3}$} &  \multicolumn{1}{l}{$\mathbf{78.2_{\uparrow 3.7}}$} \\
     \hline
     \multicolumn{1}{c}{(c)} & \multicolumn{1}{c}{} & \multicolumn{1}{c|}{\checkmark}& \multicolumn{1}{l}{$85.6_{\uparrow 4.5}$}  & \multicolumn{1}{l}{$77.6_{\uparrow 3.9}$} &  \multicolumn{1}{l}{$56.8_{\uparrow 4.0}$} &  \multicolumn{1}{l}{$\mathbf{78.5_{\uparrow 4.0}}$} \\
     \hline
     \multicolumn{1}{c}{(d)} & \multicolumn{1}{c}{\checkmark} & \multicolumn{1}{c|}{\checkmark} & \multicolumn{1}{l}{$86.1_{\uparrow 5.0}$}  & \multicolumn{1}{l}{$78.2_{\uparrow 4.5}$} &  \multicolumn{1}{l}{$57.7_{\uparrow 4.9}$} &  \multicolumn{1}{l}{$\mathbf{79.2_{\uparrow 4.7}}$} \\
     \hline
  \end{tabular}
  \vspace{-0.2in}
  \label{tab.ab}
\end{table}

\subsection{Ablation Study}
We investigate how each stream in our SSVD influences the overall performance, as summarized in Table \ref{tab.ab}. Note that all the variants of SSVD here are constructed over ResNet-101 for fair comparison. Besides the standard mean Average Precision (mAP) over all classes, we additionally adopt the metric protocols in \cite{zhu2017fgfa} by categorizing ground-truth objects with respect to their motion speed. The speed of an object is measured by averaged intersection-over-union (IoU) with its corresponding instances in the nearby frames. As such, the objects are classified into three classes: slow (IoU $>$ 0.9), medium (IoU $\in$[0.7 ,0.9]), and fast (IoU $<$ 0.7).

\textbf{Method (a)} is the single-frame baseline. By directly linking class/box subnets with FPN, (a) operates object detection over each single frame in one-stage paradigm and the mAP performance achieves 74.5\%. When detailing the performances for objects with different motion speeds, we can clearly observe that the faster the object moves, the lower the detection performance. The results indicate the challenge of object detection in videos, especially when the frames are with fast-moving objects and may be deteriorated by motion blur.

\begin{figure}
  \centering
  \includegraphics[width=1\linewidth]{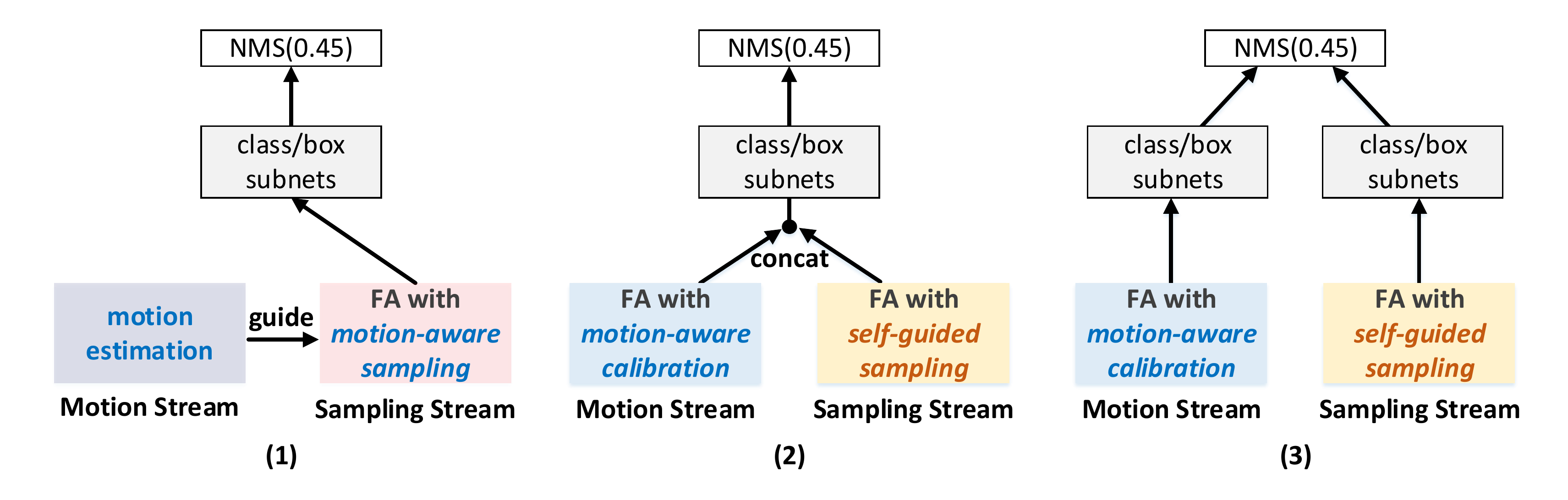}\\
  \vspace{-0.2in}
  \caption{Different schemes for fusing two streams in SSVD: (1) Integrate estimated motion from motion stream into sampling stream for feature aggregation; (2) Early fusion before class/box subnets via concatenation and (3) Late fusion with NMS. The jaccard overlap of NMS is set as 0.45.} \label{fig.fusion}
  \vspace{-0.20in}
\end{figure}

\textbf{Method (b)} integrates the feature aggregation module with motion-aware calibration into (a). The mAP increases from 74.5\% to 78.2\% and the performance improvement is consistently observed on objects in different speed. This validates the effectiveness of enhancing per-frame feature by aggregating features from nearby frames with the guidance of optical flow along motion stream.

\textbf{Method (c)} equips (a) with the feature aggregation module with self-guided sampling. (c) exhibits better performance than (a), which demonstrates the advantage of directly hallucinating features through sampling features from adjacent frames in sampling stream. Furthermore, (b) yields inferior performance to (c). The result basically indicates that relying on optical flow for feature aggregation in motion stream might more easily suffer from robustness problem (\emph{e.g.}, motion blur or occlusion) than self-learning of offset for feature hallucination in sampling stream.

\textbf{Method (d)} utilizes both motion and sampling streams with late fusion scheme, which further boosts up the performances. The results verify the merit of simultaneously exploiting feature aggregation with motion-aware calibration and self-guided sampling in one-stage paradigm.

\begin{figure}
  \centering
  \vspace{-0.1in}
  \includegraphics[width=0.9\linewidth]{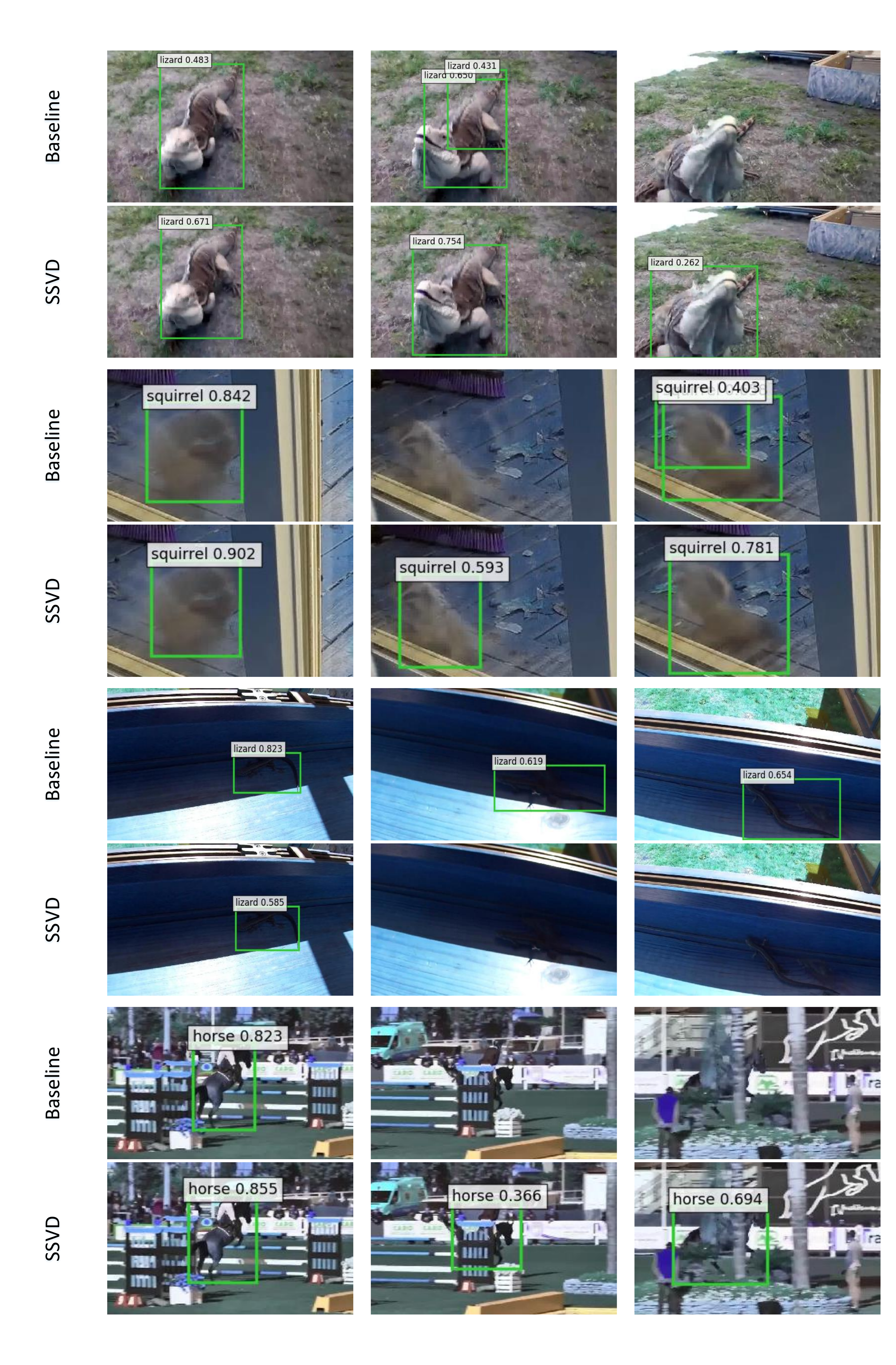}\\
  \vspace{-0.25in}
  \caption{Four video examples of detection results on ImageNet VID validation set by the single-frame baseline and our SSVD. The examples include four typical deteriorated object appearances in videos, \emph{i.e.}, rare poses, motion blur, illumination variation and part occlusion.}
  \label{fig.example}
  \vspace{-0.25in}
\end{figure}

\textbf{Effect of Fusion Scheme.} In general, there are three directions to fuse motion stream and sampling stream in our model. One is to integrate optical flow guidance into sampling stream for sampling coordinates prediction. Another is to perform early fusion scheme by concatenating aggregated features of two streams in each scale before feeding them into class/box subnets. And the last one is our adopted late fusion scheme to additionally perform NMS over all the predicted bounding boxes from two streams. Figure \ref{fig.fusion} depicts these three fusion schemes. We compare the performances of our SSVD with these three schemes. The results are 78.7$\%$, 78.1$\%$ and $79.2\%$  w.r.t motion-guided sampling, early fusion with concatenation and late fusion by NMS, which indicate that the adopted late fusion scheme outperforms the other two schemes in our case.

\textbf{Qualitative Analysis.} Figure \ref{fig.example} showcases four video examples of detection results on ImageNet VID validation set by the single-frame baseline (a) and SSVD. Four typical deteriorated object appearances in videos, \emph{i.e.}, rare poses, motion blur, illumination variation and part occlusion, are depicted in Figure \ref{fig.example} and our SSVD consistently shows better detection results than the single-frame baseline (a). For example, the single-frame baseline (a) fails to detect ``horse" in the second and third frames of the fourth video due to part occlusion. In contrast, by leveraging feature aggregation with motion-aware calibration and self-guided sampling, SSVD can detect objects in each frame correctly.

\begin{figure}
  \centering
  \vspace{-0.0in}
  \includegraphics[width=1.0\linewidth]{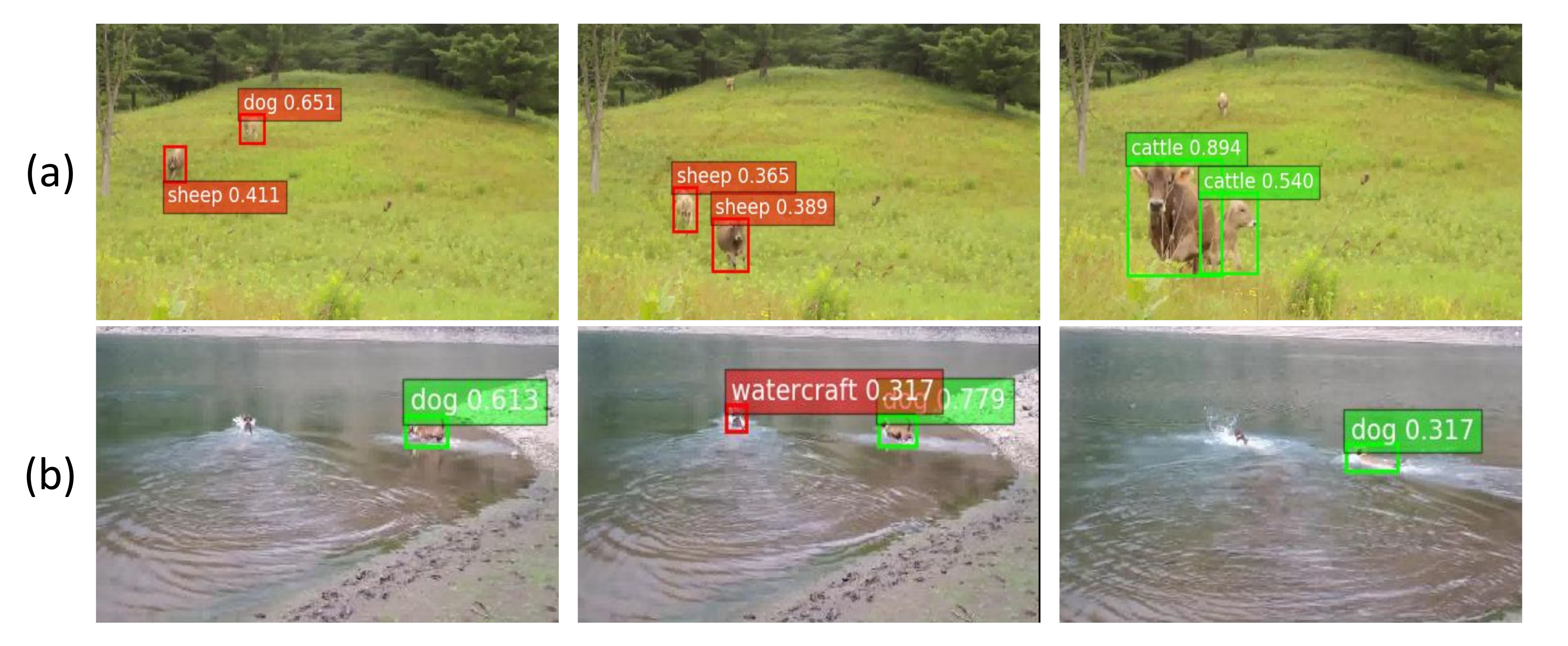}\\
  \vspace{-0.2in}
  \caption{Failure cases in the ImageNet VID dataset (better viewed in color). The green boxes illustrate the true-positive detections, and the red one shows the prediction with wrong classes.}
  \label{fig.failure}
  \vspace{-0.18in}
\end{figure}

\textbf{Failure Cases Analysis.} Here we further showcase some failure cases of our SSVD. As shown in Figure \ref{fig.failure}, for case (a), our SSVD fails to correctly detect smaller objects, i.e., two cattle, when they are far away from the camera. Nevertheless, after a long range of time, our SSVD succeeds to detect them when they are near the camera. This problem can be alleviated by completing the detection results of smaller objects with more accurate ones when they are near camera. Therefore, how to preserve such long-range temporal coherence might strengthen our SSVD, which will be one of our future works. Moreover, in case (b), for the two dogs in each frame, we can easily detect one in the right, while our SSVD fails to correctly localize or recognize the left one. We speculate that this failure case is due to that SSVD leaves the relations between objects in each frame unexploited. One possible way to address it is to enable the interactions between objects in SSVD.

\begin{table}[!tb]
  \centering
  \footnotesize
  \caption{Performance comparisons with the state-of-the-art end-to-end video object detection methods on ImageNet VID validation set. The mean average precision (mAP) over all classes is reported.}
  \begin{tabular}{l|c|c}
   \Xhline{2\arrayrulewidth}
  Methods                           & Backbone                                            & mAP (\%)      \\
  \hline\hline
  R-FCN\,\cite{dai2016r}   & ResNet-101 & 73.6 \\
  \hline
  DorT\,\cite{luo2019detect}  & ResNet-101 & 73.9 \\
  \hline
  Faster R-CNN\,\cite{ren2015faster}  & ResNet-101 & 75.4 \\
  \hline
  D\,(\& T\,loss)\,\cite{feichtenhofer2017detect}  & ResNet-101 & 75.8 \\
  \hline
  FGFA\,\cite{zhu2017fgfa}  &ResNet-101                     & 76.3          \\
  \hline
  LWDN\,\cite{jiang2019video} & ResNet-101 & 76.3 \\
  \hline
  PSLA\,\cite{guo2019progressive}  & ResNet-101 &  77.1\\
  \hline
  MANet\,\cite{wang2018manet}       & ResNet-101                                                 & 78.1          \\
  \hline
  THP\,\cite{zhu2018towards}        & ResNet-101+DCN            & 78.6          \\
  \hline
  STSN\,\cite{stsn}                 & ResNet-101+DCN        & 78.9          \\

  \Xhline{2\arrayrulewidth}
  \multirow{3}{*}{SSVD}            & ResNet-101                                         &$\mathbf{79.2}$ \\
                                                 & ResNet-101+DCN  & $\mathbf{80.2}$\\
                                                 & ResNeXt-101-32$\times$4d                           &$\mathbf{81.8}$ \\
  \Xhline{2\arrayrulewidth}
  \end{tabular}
  \vspace{-0.15in}
  \label{tab.vid_val}
  \end{table}

\subsection{Comparisons with State-of-the-Art Methods}
\textbf{End-to-end Models.} Table \ref{tab.vid_val} compares the performances of nine runs on ImageNet VID validation set, including several state-of-the-art techniques and our SSVD. These models purely learn video object detector by enhancing per-frame feature for detection in an end-to-end fashion without any post-processing. Among them, the former two approaches (\emph{i.e.}, R-FCN \cite{dai2016r} and Faster R-CNN \cite{ren2015faster}) are two representative two-stage image object detectors which only exploit single frame information. D (\&T loss) \cite{feichtenhofer2017detect} exhibits better performances than the two image object detectors by exploiting temporal coherence among adjacent frames via RoI tracking module. The latter three baselines (\emph{i.e.}, FGFA \cite{zhu2017fgfa}, MANet \cite{wang2018manet}, and STSN \cite{stsn}) further boost the performance for video object detection by enhancing per-frame feature. Specifically, FGFA takes optical flow as guidance for pixel-level calibration for feature aggregation. MANet outperforms FGFA by additionally utilizing box-level calibration for feature aggregation. STSN leverages spatio-temporal sampling instead of motion calibration for feature aggregation, which achieves comparable performance with FGFA under the same backbone of Deformable Convolution Network \cite{dai2017deformable} (DCN). Note that as reported in \cite{zhu2018towards}, the performance of FGFA with DCN is 78.8\%. For fair comparisons, our SSVD is evaluated based on three commonly adopted basic architectures, ResNet-101, Deformable ResNet-101 (DCN), and ResNeXt-101-32$\times$4d \cite{xie2017aggregated} . Overall, the results with the same basic architecture consistently demonstrate that our proposed SSVD by integrating two-stream feature aggregation into one-stage detection paradigm exhibits better performance than all the six two-stage end-to-end models.

\begin{table}[!tb]
  \centering
  \footnotesize
  \caption{Performance comparisons with the state-of-the-art video object detection systems plus post-processing on ImageNet VID validation set. The mean average precision (mAP) over all classes is reported. $^\triangle$ indicates linking with Seq-NMS \cite{han2016seq}.}
  \begin{tabular}{l|c|c}
   \Xhline{2\arrayrulewidth}
  Methods                           & Backbone                                            & mAP (\%)      \\
  \hline\hline
  TPN+LSTM \cite{kang2017object} & GoogLeNet & 68.4 \\
  TCNN\,\cite{kang2017t}  & DeepID+Craft & 73.8  \\
  FGFA\,$^\triangle$\,\cite{zhu2017fgfa} & ResNet-101 & 78.4 \\
  D\&T\,($\tau=1$) \cite{feichtenhofer2017detect}  & ResNet-101 & 79.8 \\
  STSN\,$^\triangle$\,\cite{stsn}  & ResNet-101+DCN & 80.4 \\
   STMN \cite{xiao2018matchtrans}  & ResNet-101 & 80.5 \\
    HQ-link\,\cite{tang2019object} & ResNet-101 & 80.6 \\
  \Xhline{2\arrayrulewidth}
  \multirow{3}{*}{SSVD\,$^\triangle$}            & ResNet-101                                         &$\mathbf{80.5}$ \\
                                                 & ResNet-101+DCN  & $\mathbf{81.1}$\\
                                                 & ResNeXt-101-32$\times$4d                           &$\mathbf{82.8}$ \\
  \Xhline{2\arrayrulewidth}
  \end{tabular}
  \vspace{-0.15in}
  \label{tab.comparison_plus_post}
  \end{table}

\textbf{Post-processing.} Furthermore, we equipped our SSVD with Seq-NMS \cite{han2016seq}, a larger degree of improvement is attained (\emph{e.g.}, from 79.2\% to 80.5\% for ResNet-101), achieving competitive results with state-of-the-art video object detectors plus the post-processing of box-level association. The comparison results are illustrated in Table \ref{tab.comparison_plus_post}. Technically, both of FGFA and STSN adopt Seq-NMS as post-processing. TPN+LSTM exploits encoder-decoder LSTM to rescore generated tubelets proposal by Tubelet Proposal Networks (TPN) \cite{kang2017object} and the performance is 68.4\%. TCNN performs tubelet rescoring by motion-guided propagation to stabilize detection results and achieves 73.8\% mAP. D\&T links detection boxes between each two adjacent frames by predicted tracking boxes and Viterbi Algorithm as in \cite{gkioxari2015finding}, which boosts the performance from 75.8\% to 79.8\%. STMN firstly combines detection results of spatio-temporal memory module based sequence prediction with that of single-frame baseline, and then takes similar linking algorithm as D\&T. In addition, the performance of our SSVD with post-processing is even comparable to the winner of ILSVRC2016 \cite{ilsvrc16NUIST}, which is a comprehensive detection system. Specifically, by utilizing multi-scale testing and model ensemble, \cite{ilsvrc16NUIST} achieves 81.2\% mAP, which is similar to the result of our SSVD with ResNeXt-101-32$\times4$d.

\begin{table}[!tb]
\centering
  \footnotesize
  \setlength{\tabcolsep}{3.2pt}
  \caption{The effect of the number of sampled support frames in our SSVD at inference on ImageNet VID dataset. The experiments are conducted on an Nvidia Titan X Pascal GPU.}
  \begin{tabular}{c|c|c|c|c|c|c}
    \hline
    \# sampled support frames  & 0  & 2  & 4     &  6    &  12    &  24   \\
    \hline
    mAP (\%)     & 74.52 & 77.1 & 79.07  &  79.18 &  79.12  & 79.16  \\
    \hline
    run time (ms)  & 38 & 66 & 74   &  85 &  114   & 183   \\\hline
  \end{tabular}
  \label{tab.supportnum}
  \vspace{-0.2in}
\end{table}

\begin{figure}
  \centering
  \vspace{-0.2in}
  \includegraphics[width=0.9\linewidth]{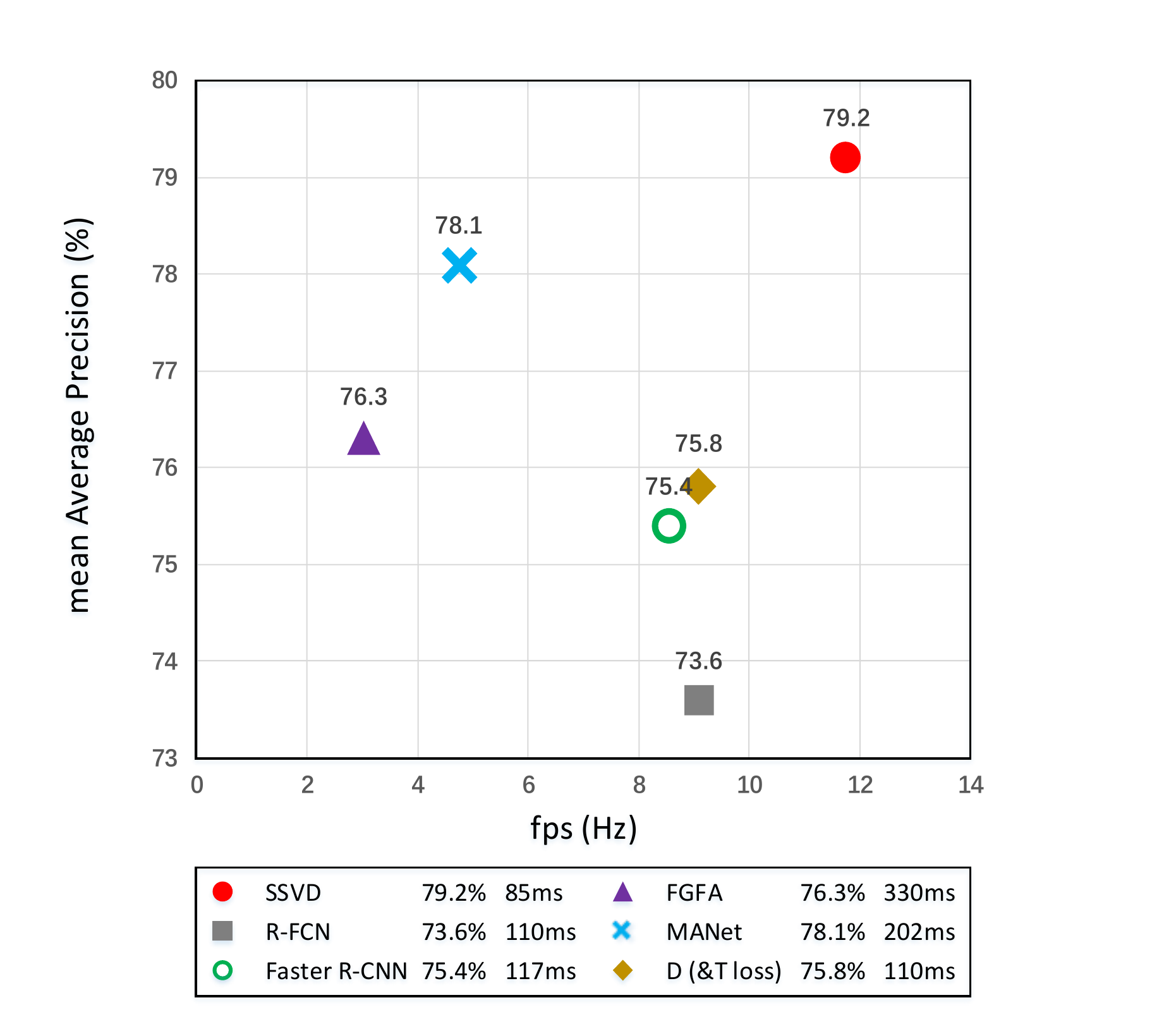}\\
  \vspace{-0.2in}
  \caption{Performance and run time comparisons with the state-of-the-art methods at inference on ImageNet VID dataset. The mean Average Precision and average runtime for each method are detailed in the legend below.}
  \label{fig.runtime}
  \vspace{-0.22in}
\end{figure}

\subsection{Experimental Analysis}\label{sec.runtime}
\textbf{Effect of the Number of Sampled Support Frames.} As mentioned in implementation details, we uniformly select support frames from the sliding feature buffer (25 adjacent frames) at inference stage. In order to show the relationship between the performance/run time and the number of sampled support frames, we compare the results by varying this number from 0 to 24. As shown in Table \ref{tab.supportnum}, except for the setting with 0 sampled support frame (\emph{i.e.}, our single-frame baseline), the performance by using different number of sampled support frames fluctuates within the range of 0.11\%. That practically eases the option of the number of sampled support frames. Meanwhile, enlarging the number of sampled support frames generally increases run time at inference stage. Therefore, in our experiments, the number of sampled support frames is empirically set to 6, which is a good tradeoff between performance and run~time.

\textbf{Run Time Comparison.} Figure \ref{fig.runtime} depicts the detailed run time and performance of each approach. The results clearly indicate the advantage of SSVD against state-of-the-art two-stage approaches in terms of both speed and accuracy. In particular, at inference stage, SSVD processes one frame in 85 ms, which is even faster than the single-frame baseline of two-stage video object detectors, \emph{e.g.}, R-FCN and Faster R-CNN. Note that here we exclude several video object detection methods (\cite{zhu2018towards,zhu2017dff}) which are additionally equipped with the acceleration techniques (\emph{e.g.}, flow guided feature propagation and adaptive key frame selection). The similar acceleration techniques can also be adopted in our SSVD to further enhance the efficiency, and we will leave this discussion for further investigation.

\textbf{Tracklet Evaluation.} We additionally evaluate our SSVD over an auxiliary tracking metric (mAP$^{track}$) in ImageNet VID. The mAP$^{track}$ metric measures the mAP of predicted tracklet against ground truth tracklet in a snippet, which goes beyond the mAP metric that only evaluates the detection accuracy at frame level. Specifically, we exploit the Viterbi algorithm to link the per-frame detection boxes of our SSVD into tracklets. The mAP$^{track}$ of SSVD + Viterbi \cite{gkioxari2015finding} is 60.2\%, which is higher than 57.0\% of DorT \cite{luo2019detect}. The results further demonstrate the effectiveness of our SSVD at tracklet level.

\textbf{Take-aways. } Here we summarize several take-aways as follows: (1) Motion stream often fails to estimate motion from optical flow when the object appearances are deteriorated by motion blur or occlusion, while sampling stream can hallucinate the feature map via frame sampling and thus performs better in this case. (2) In contrast, sampling stream may suffer from robustness problem when object moves extremely fast. This is due to the fact that the receptive field in sampling stream for offset prediction is smaller than that in PWC-Net \cite{sun2018pwc} for optical flow generation. As such, the range of estimated motion in sampling stream is shorter than that in motion stream, resulting in failure of motion capturing in sampling stream when object moves extremely fast. In this case, motion stream performs better than sampling stream. (3) In practice, considering the information redundancy across adjacent frames, we uniformly sample 6 support frames from the sliding feature buffer (25 adjacent frames) to perform detection at inference. As shown in Table \ref{tab.supportnum}, this selected number of sampled support frames indeed seeks good tradeoff between performance and run time.

\section{Conclusions}
We have presented Single Shot Video Object Detector (SSVD), which explores temporal coherence across frames to boost video object detection. Particularly, we study the problem from the viewpoint of integrating two-stream feature aggregation into a single shot detector. To verify our claim, we utilize FPN to produce multi-scale feature maps in a spatial pyramid and feed feature maps in each scale into two-stream feature aggregation structure. One is motion stream that performs feature aggregation by estimating motion from optical flow and warping the feature maps of nearby frames to the reference one along motion paths. The other is sampling stream, which directly hallucinates the feature map of the reference frame through spatio-temporal sampling from adjacent frames. The aggregated feature map in each stream is input into class/box subnets to simultaneously classify anchor boxes and perform bounding box regression. Extensive experiments conducted on ImageNet VID dataset validate our proposal and analysis. More remarkably, we achieve state-of-the-art performance of single model without post-processing: 79.2\% mAP, by processing one frame in 85 ms on an Nvidia Titan X Pascal~GPU.

{\small
\bibliographystyle{IEEEtran}
\bibliography{egbib}
}

\end{document}